\documentclass[a4paper,twoside]{article}

\usepackage{epsfig}
\usepackage{subcaption}
\usepackage{calc}
\usepackage{amssymb}
\usepackage{amstext}
\usepackage{amsmath}
\usepackage{amsthm}
\usepackage{multicol}
\usepackage{pslatex}
\usepackage{apalike}
\usepackage{algorithm2e}
\usepackage{tikz}
\usepackage{graphicx}
\usepackage{subcaption}

\usetikzlibrary{positioning, arrows.meta, fit}

\usepackage[bottom]{footmisc}
\usepackage{SCITEPRESS}     

\usepackage{soul,xcolor}

\begin{document}

\title{Enhancing Manufacturing Quality Prediction Models through the Integration of Explainability Methods}

\author{\authorname{Dennis Gross\sup{1}, Helge Spieker\sup{1}, Arnaud Gotlieb\sup{1}, and Ricardo Knoblauch\sup{2}}
\affiliation{\sup{1}Simula Research Laboratory, Oslo, Norway}
\affiliation{\sup{2}Ecole Nationale Supérieure d'Arts et Métiers, Aix-en-Provence, France}
\email{{dennis@simula.no}
}}


\abstract{This research presents a method that utilizes explainability techniques to amplify the performance of machine learning (ML) models in forecasting the quality of milling processes, as demonstrated in this paper through a manufacturing use case. The methodology entails the initial training of ML models, followed by a fine-tuning phase where irrelevant features identified through explainability methods are eliminated. This procedural refinement results in performance enhancements, paving the way for potential reductions in manufacturing costs and a better understanding of the trained ML models. This study highlights the usefulness of explainability techniques in both explaining and optimizing predictive models in the manufacturing realm.}
\onecolumn \maketitle \normalsize \setcounter{footnote}{0} \vfill

\section{\uppercase{Introduction}}
\label{sec:introduction}
\emph{Milling} is a subtractive manufacturing process that involves the removal of material from a workpiece to produce a desired shape and surface finish.
In this process, a cutting tool known as a milling cutter rotates at high speed and moves through the workpiece, removing material.
The workpiece is typically secured on a table that can move along multiple axes, allowing for various orientations and angles to be achieved~\cite{fertig2022machine}.
The energy consumption during the milling process can vary significantly based on several factors, including the specific setup and the materials being processed.
However, it is often considered to be a relatively energy-intensive process.
Being able to foresee and avert potential quality issues means that less energy is expended and fewer resources are wasted on creating defective parts, which would otherwise be rejected or require reworking~\cite{pawar2021modelling}.

\emph{Machine learning (ML) models} are capable of identifying patterns and structures in data to make predictions without being directly programmed to do so.
These models can be a useful tool in forecasting the final quality of a workpiece at the end of the milling process, helping to improve both the efficiency and the reliability of the manufacturing process~\cite{mundada2018optimization}.

\paragraph{Lack of data.}
However, the available milling experiment data is typically small because the costs for each manufacturing experiment are very high~\cite{postel2020ensemble}.
This lack of data makes it difficult to learn ML models to predict the workpiece quality.

\paragraph{Explainability problem.}
Even if abundant data were available, utilizing complex ML models, such as deep neural network models, presents a significant challenge due to their ``black box'' nature.
This term denotes the lack of transparency in understanding these models' internal workings, which often remain inaccessible or unclear~\cite{DBLP:journals/cogsr/KwonKC23}.
In the realm of milling process quality prediction, an explainability problem arises when practitioners and stakeholders can not fully understand the predictions given by these models due to their complex and opaque nature.

\paragraph{Leveraging explainability techniques for optimized model training.}
Explainability methods are, therefore, crucial in unraveling the complex prediction mechanisms embedded within ML models.
Furthermore, they facilitate enhanced performance of ML models by identifying and addressing potential inefficiencies, thereby steering optimization efforts more effectively~\cite{bento2021improving,sun2022utilizing}.

\paragraph{Approach.}
We use less computational processing-intense ML models, such as decision tree regression or gradient boosting regression, and improve their performance via explainability methods.
We do this by training these models on a milling dataset, identifying the important features for these models for their predictions, removing the less important features from them, and retraining a new model on stretch on the feature-pruned data set.
Our results show, on a case study from the manufacturing industry in the context of milling, that we can improve the performance of ML models and reduce financial manufacturing costs by pruning the features.
These findings indicate that the strategies used in recent studies for enhancing ML model performance through explainability~\cite{bento2021improving,sun2022utilizing} can be effectively adapted for manufacturing processes.

\paragraph{Plan of the paper.} The rest of the paper is organized as follows: Section~\ref{sec:related} covers existing works on using AI for manufacturing/machining problems; Section~\ref{sec:metho} presents our explainability methodology and how to apply it to ML models; Section~\ref{sec:case} explains how to deploy our explainability methodology to quality prediction model used in surface milling operations.
Eventually, Section~\ref{sec:discu} and Section~\ref{sec:conclu} respectively discuss the benefits and drawbacks of using explainability methods in machining and conclude the paper by drawing some perspectives to this work.

\section{Related Work}
\label{sec:related}

The usage of ML in manufacturing/machining tasks has been recognized as an interesting lead for at least a decade \cite{Kummar17}. For instance, ML has been used initially to optimize turning processes \cite{mokhtari2014optimization}, predicting stability conditions in milling \cite{postel2020ensemble}, estimating the quality of bores \cite{schorr20}, or classifying defects using ML-driven surface quality control \cite{Chouhad21}.

However, it is only recently that Explainable AI (XAI) methods have been identified as an interesting approach for manufacturing processes \cite{Yoo21,DBLP:journals/mansci/SenonerNF22}.
The ongoing European XMANAI project \cite{Lampathaki21} aims to evaluate the capabilities of XAI in different sectors of manufacturing through the development of several use cases.
In particular, fault diagnosis seems to be an area where XAI can be successfully applied \cite{Brusa23}. Also, there exists work that focuses on feature selection on the dataset without taking the ML model directly into account~\cite{bins2001feature,oreski2017effects,venkatesh2019review}. Our approach in this paper is to explore the potential of XAI for enhancing quality prediction models by eliminating unnecessary sensors. Even though the approach of improving ML models via explainability methods is known in the context of explainable ML~\cite{bento2021improving,sun2022utilizing,DBLP:conf/ilp/0002S21,sofianidis2021review}, to the best of our knowledge, it is the first time that XAI is used for that application. In particular, identifying unnecessary features through XAI methods to improve quality prediction models in milling processes is novel.

\section{\uppercase{Methodology}}
\label{sec:metho}
Our approach works as follows:
First, we train an ML model on the given data set.
Second, we apply an explainability method to the ML model and the data set to identify the most important features for the prediction accuracy.
Third, we rank the features based on their feature importance in a descending ordered sequence and incrementally increase the number of features being used for the training (for each new setting a new ML training).
Fourth, we take the features that led to the best-performing model.
We now explain the different steps in more detail.

\subsection{Machine Learning Models}
In this study, we use decision tree regression, gradient boosting regression, and random tree regression models for the prediction process. These ML models are less data-intensive than neural networks and are considered easier to interpret.

\paragraph{Decision Tree Regression Models.} 
Decision tree regression models partition the input space into distinct regions and fit a simple model (typically a constant) to the training samples in each region. For a new input \( x \), the prediction \( \hat{y} \) is obtained from the model associated with the region \( R_m \) that \( x \) belongs to.
The prediction can be expressed formally as:

\[
\hat{y}(x) = \sum_{m=1}^{M} c_m I\{x \in R_m\}
\]

where \( c_m \) represents the constant fitted to the samples in region \( R_m \), \( M \) is the number of regions, and \( I\{\cdot\} \) is an indicator function. This model is capable of capturing non-linear relationships through a piecewise linear approach~\cite{myles2004introduction}.

\paragraph{Gradient Boosting Regression Models.} 
Gradient boosting regression models optimize a given loss function \( L(y, \hat{y}(x)) \) by combining multiple weak models. The model starts with an initial approximation \( F_0(x) \) and iteratively refines this through the addition of weak models \( h_m(x) \). The update at each iteration \( m \) is described by:

\[
F_m(x) = F_{m-1}(x) + \alpha \cdot h_m(x)
\]
where \( F_m(x) \) is the model at iteration \( m \), \( \alpha \) is the learning rate, and \( h_m(x) \) is a weak learner aimed at correcting the errors of the previous models. This iterative process results in a robust predictive model~\cite{otchere2022application}.

\paragraph{Random Tree Regression Models.} 
Random Tree Regression models also employ an ensemble learning strategy, building multiple decision trees during the training phase and aggregating them for predictions. The final prediction \( \hat{y} \) for an input \( x \) is the average prediction across all trees in the ensemble:

\[
\hat{y}(x) = \frac{1}{T} \sum_{t=1}^{T} y_t(x)
\]
where \( T \) is the total number of trees and \( y_t(x) \) is the prediction of the \( t \)-th tree. This aggregation helps enhance the model's generalization capabilities and mitigates the risk of overfitting~\cite{prasad2006newer}.

\subsection{Explainability Methods}

\paragraph{Feature permutation importance.}
A critical aspect of our approach is employing feature permutation importance as a significant explainability method. This technique operates by evaluating the importance of different features in the model. The general procedure involves the random permutation of a single feature, keeping others constant, and monitoring the change in the model's performance, often measured through metrics like accuracy or mean squared error~\cite{huang2016permutation}.

Mathematically, the feature importance \( I_i \) of a feature \( i \) can be defined as the difference in the model's performance before and after the permutation of the feature and can be formulated as:
\[ I_i = P_{original} - P_{permuted(i)} \]
where \( P_{original} \) is the model's performance with the original data and \( P_{permuted(i)} \) is the performance with the \(i\)-th feature permuted.

By iterating this process across all features and comparing the changes in performance, we can rank the features by their importance, offering deeper insights into the model's decision-making process and enabling the identification of areas for optimization and refinement.

\paragraph{Shapley values.}
The Shapley value, originating from cooperative game theory, allocates a fair contribution value to each participant based on their marginal contributions to the collective outcome. In the context of machine learning models, Shapley values are used to quantify the contribution of each feature to the prediction made by the model~\cite{sundararajan2020many}.

Given a predictive model \( f: \mathbb{R}^n \rightarrow \mathbb{R} \), where \( n \) is the number of features, the Shapley value of the \( i \)-th feature is calculated as follows:

\[
\phi_i(f) = \sum_{S \subseteq N \setminus \{i\}} \frac{|S|! \cdot (n - |S| - 1)!}{n!} \left[ f(S \cup \{i\}) - f(S) \right]
\]

where:
\begin{itemize}
    \item \( N \) is the set of all features.
    \item \( S \) is a subset of \( N \) not containing feature \( i \).
    \item \( |S| \) denotes the number of features in subset \( S \).
    \item \( f(S) \) is the prediction of the model when only the features in \( S \) are used.
    \item The term \( f(S \cup \{i\}) - f(S) \) represents the marginal contribution of feature \( i \) when added to the subset \( S \).
\end{itemize}

The Shapley value \( \phi_i(f) \) captures the average contribution of feature \( i \) to the model's prediction, averaged over all possible subsets of features. This comprehensive averaging process ensures a fair distribution of contributions, considering all possible interactions between features.

It is important to note that the computation of Shapley values can be particularly intensive, especially in scenarios where the number of features is large. This is due to the necessity to evaluate the model's prediction for every possible combination of features, resulting in a total of \( 2^n \) model evaluations for \( n \) features. Therefore, practical implementations often employ approximation methods or sampling strategies to mitigate the computational demands.

\section{\uppercase{Case Study}}
\label{sec:case}
In this case study, we apply our method to a dataset generated at MSMP - ENSAM. A series of surface milling operations were performed on aluminum 2017A using a 20 mm diameter milling cutter R217.69-1020.RE-12-2AN with two carbide inserts XOEX120408FR-E06 H15 from SECO, and a synthetic emulsion of water and 5\% of Ecocool CS+ cutting fluid. 

In total, 100 experiments have been carried out varying the following process parameters: depth of cut, cutting speed, and feed rate.
For each one of these experiments with different control parameters, cutting forces $F_z$ (normal force) and $F_a$ (active force), and surface profiles are measured on-machine using a Kistler 3-axis dynamometer 9257A and a STIL CL1-MG210 chromatic confocal sensor (non-contact) respectively.
In the feature engineering step, the following surface roughness amplitude parameters are calculated using MountainsMap software. Ra is the most commonly used in the industry.
\begin{itemize}
    \item Ra (Average Roughness): Average value of the absolute distances from the mean line to the roughness profile within the evaluation length. 
    \item Rz (Average Maximum Height): Average value of the five highest peaks and the five deepest valleys within the evaluation length. 
    \item Rt (Total Roughness): Vertical distance between the highest peak and the deepest valley within the evaluation length. 
    \item Rq (Root Mean Square Roughness): Square root of the average of the squared distances from the mean line to the roughness profile within the evaluation length.
    \item RSm (Mean Summit Height): Average height of the five highest peaks within the evaluation length. 
    \item RSk (Skewness): Measure of the asymmetry of the roughness profile around the mean line. 
    \item Rku (Kurtosis): Measure of the peakedness or flatness of the roughness profile. 
    \item Rmr (Material Ratio): Ratio of the actual roughness profile area to the area within the evaluation length. 
    \item Rpk (Peak Height): Height of the highest peak within the evaluation length. 
    \item Rvk (Valley Depth): Depth of the deepest valley within the evaluation length. 
    \item Rdq: It is a hybrid parameter (height and length). It is the root mean square slope of the assessed profile, defined on the sampling length. Rdq is the first approach to surface complexity. A low value is found on smooth surfaces, while higher values are found on rough surfaces with microroughness.
\end{itemize}

\begin{figure}[t]
    \centering
    \includegraphics[width=0.4\textwidth]{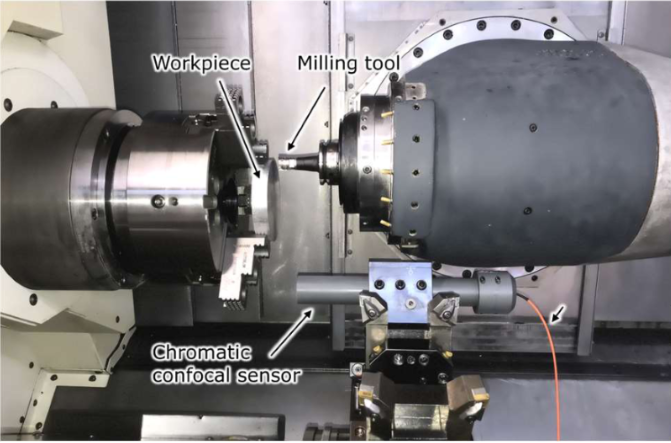}
    \caption{Milling machine that produces workpieces.}
    \label{fig:machine1}
\end{figure}

\subsection{Objective}
The aim is to develop a predictive model for each quality metric associated with roughness amplitude parameters.
This necessitates not only the training of accurate models but also an elucidation of the predictive rationales behind their outputs.
Concurrently, there is a need to identify and eliminate superfluous features from the models.
This is a strategic step to minimize both installation and maintenance expenses related to redundant sensors, thereby optimizing resource allocation and reducing overall costs.

\subsection{Data Preprocessing}
Since we are dealing with variable time series lengths, we calculate the box plot values for each time series in the time and frequency domain.
Additionally, metadata within the dataset comprises experiment parameters with various focuses.

\subsection{ML Model Training}
\begin{figure}[]
    \begin{tikzpicture}[>=Stealth, node distance=0.25cm]
        \node (input1) {$F_A$ Box Plot};
        \node [below=0.1cm of input1] (input2) {$F_Z$ Box Plot};
        \node [below=0.1cm of input2] (input3) {Machine Config};
        
        \node [right=0.5cm of input2, draw, rounded corners, inner sep=20pt] (model) {Model};

        \node [right=1cm of model] (output) {Quality};
    
        \draw[->] (input1) -- (model);
        \draw[->] (input2) -- (model);
        \draw[->] (input3) -- (model);
        
        \draw[->] (model) -- (output);
    \end{tikzpicture}
    \caption{The ML prediction model receives the box plots (for time and frequency domains) and machine configuration parameters to output the quality measures.}
    \label{fig:prediction_model}
\end{figure}
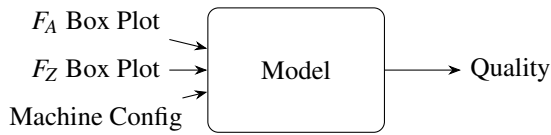

We trained a decision tree regression, gradient boosting regression, and random forest models for each quality measure.
See the overall input-output of the models in Figure~\ref{fig:prediction_model}.
We employ a 5-fold cross-validation approach for each model training. In this method, the data is divided into five equal parts. In each iteration, four parts (80\%) are used for training the model, and the remaining one part (20\%) is utilized for testing. This process is repeated five times, with each of the five parts serving as the test set exactly once. The performance of the model is then averaged over the five iterations to obtain a more robust estimate of its effectiveness.

\subsection{Analysis}
In this section, we analyze our approach.
First, we evaluate the ML model performance.
Second, we analyze the predictive mechanisms of the ML models.
Third, we evaluate the effect on the ML model performances by removing features.

\begin{figure*}[t]
    \centering
    \includegraphics[width=1\textwidth]{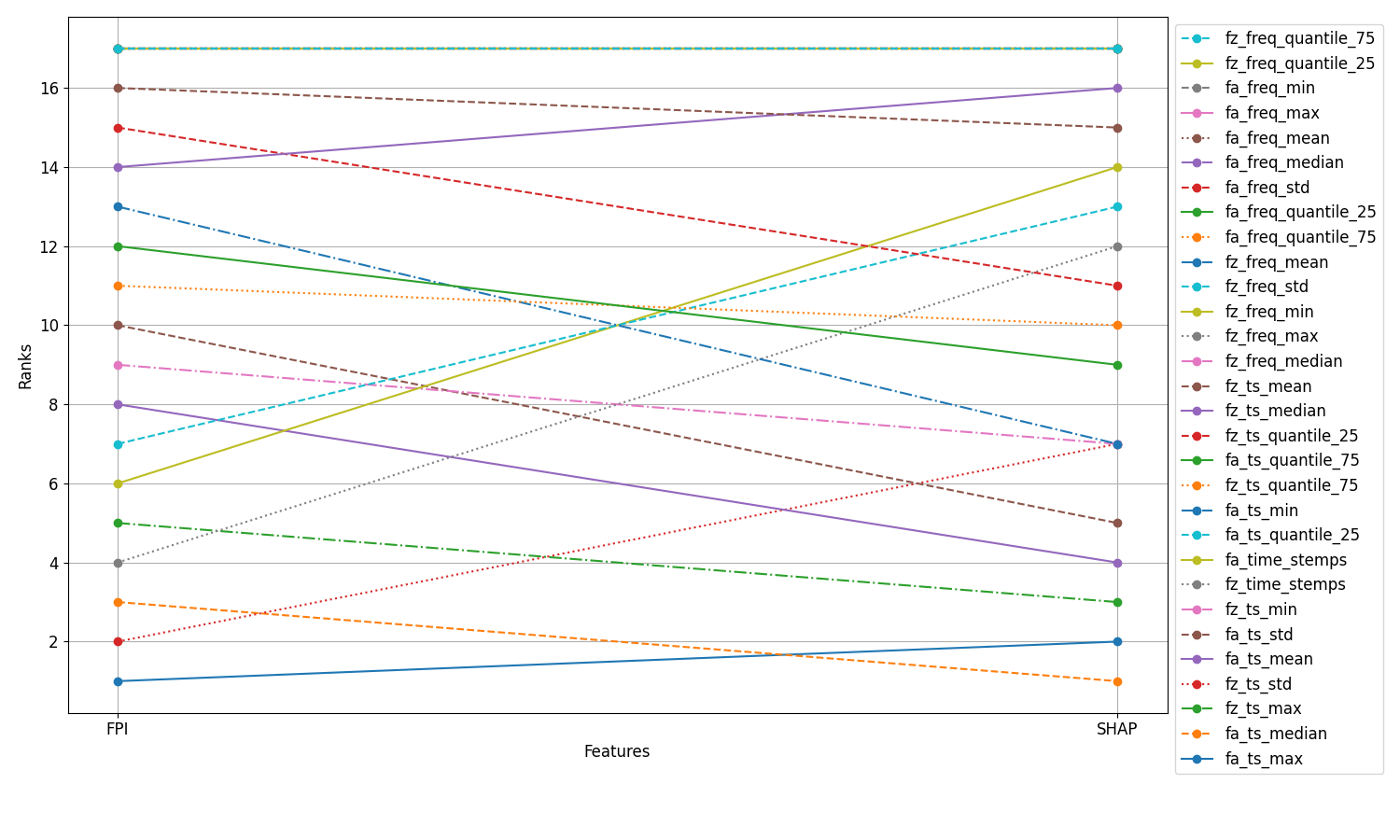}
    \caption{Visualization demonstrating the feature importance rank of Rdqmaxmean predictions, as elucidated by the feature permutation importance permutation (FPI) and Shapley value (SHAP) method.}
    \label{fig:both}
\end{figure*}

\subsubsection{Evaluating the Predictive Quality of the Models}
The objective of this study is to evaluate the predictive accuracy of three distinct ML models: gradient boosting regression, decision tree, and random forest. In assessing the quality of the predictions, the key metric employed is the \emph{Mean Absolute Percentage Error (MAPE)}. A prediction is considered to be of high quality if its MAPE is below 5\%.

\paragraph{Setup.} Preprocessed dataset with 100 samples.

\paragraph{Execution.}
We trained three different machine learning models: gradient boosting regression, decision tree, and random forest on the preprocessed dataset and used k-fold cross-validation to measure their MAPE.

\paragraph{Findings.}
By employing these ML techniques on an exhaustive set of quality features, we achieved predictions for \textit{Rdq} with an error rate of less than $5\%$.
Specifically, the Gradient Boosting Regression model yielded an error rate of \(4.58\%\), while the Random Forest model yielded an error rate of \(4.88\%\).

\subsubsection{Understanding the Predictive Mechanisms of the ML Models}
This study assessed the importance of various attributes in forecasting quality metrics. 

\paragraph{Setup.}
Given the trained models, we focus here on the gradient-boosting regression model, which achieved the highest performance.

\paragraph{Execution.}
We applied feature permutation importance and Shapley value to the prediction model.

\paragraph{Findings.}
We observed that the different explanation methods yield different reasons. For instance, permutation feature importance highlights fa\_ts\_max more important than the Shapley method (see rank one vs. rank two Figure~\ref{fig:both}).
In the justification of our experiment, it is crucial to acknowledge that employing various explanation methods to interpret ML models is expected to yield divergent explanations. This disparity stems from the intrinsic differences in methodological approaches, the complexity of the models, the data dependence of explanations, and the approximations and assumptions inherent in each explanation method. These factors collectively contribute to variations in identifying important features~\cite{lozano2023comparison}.

\subsubsection{Performance Improvements}\label{sec:perfimprovement}
In this experiment, we aimed to explore the potential benefits of integrating explainability methods into the ML model development process, focusing on enhancing model performance.

\paragraph{Setup.}
To initiate this, we categorized the variables in our ML model based on their respective feature importance, arranging them in descending order. This categorization enabled us to identify and quantify the significance of each feature in the context of model training.
We did this for feature permutation importance, Shap values, and a baseline feature selection method.
The baseline method is \emph{SelectKBest method}, a component of the Scikit-learn library~\cite{pedregosa2011scikit}, for the purpose of univariate feature selection. This technique operates on the principle of evaluating individual features based on statistical tests, ultimately selecting the $k$ features that demonstrate the highest relevance or significance.

\paragraph{Execution.}
Following the setup phase, we conducted a series of trials where we systematically varied the top percentage ($p$) of important features incorporated into the training dataset. In each trial, a new model was trained exclusively on the top $p\%$ of the most significant features, excluding potentially less impactful features from the model training process.

\begin{figure}[t]
    \centering
    \includegraphics[width=0.56\textwidth]{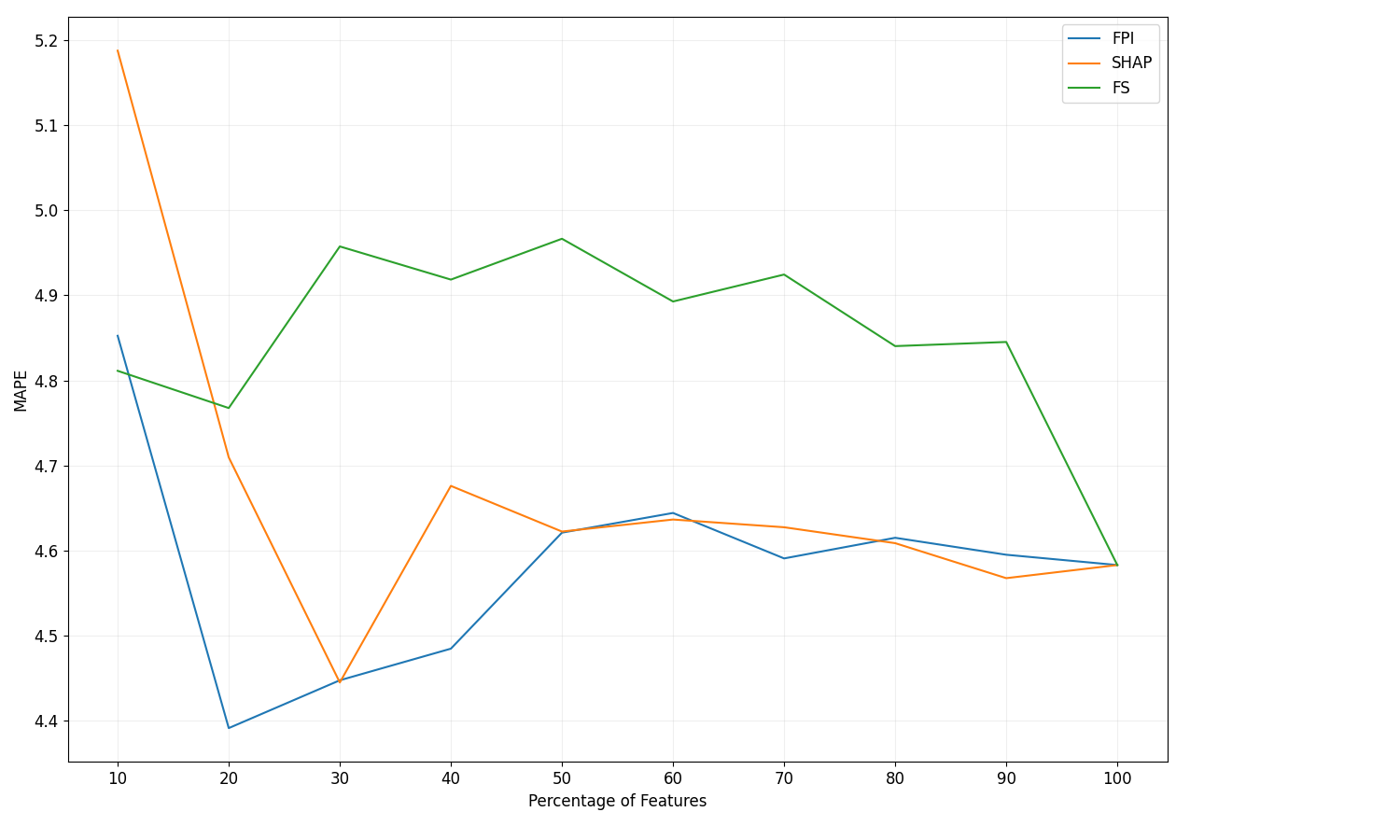}
    \caption{Using a different percentage of the most important features based on the different methods for the Rdq prediction. $FS$ refers to feature selection.}
    \label{fig:performance}
\end{figure}

\paragraph{Findings.}
We enhanced the ML models' performance by integrating only the most critical features into the training dataset (see Figure~\ref{fig:performance}).
This approach effectively streamlined the model by removing unnecessary features, improving performance, and potentially allowing for more transparent and explainable model operations.
For example, by choosing only the top 20\% of features deemed most critical (as determined by permutation importance), we enhanced the MAPE from approximately $4.58$ to $4.4$.

\section{\uppercase{Discussion}}
\label{sec:discu}
Our case study shows the benefit of explainable ML techniques for quality prediction models in manufacturing.
Explainability scores, as provided by the feature importance studied in our case study, allow to interpret the results and determine the relevance of individual features for the predictive power of a model.
This interpretation can be utilized by human domain experts for the analysis of the trained model and a plausibility check, whether those features with high importance are meaningful for the predictive task.
While it is known that ML models might reveal previously unknown relationships between input features and the prediction target, this is a rather unlikely case in our quality prediction setting.
Instead, an overreliance on a specific feature can be an indicator for learning a spurious correlation between an input and the target, potentially caused by a lack of relevant data in the small data regime. 
Explainability can, therefore, serve as an instrument for model validation and human inspection.

Furthermore, the feature importance scores support the improvement of models, as demonstrated in Section~\ref{sec:perfimprovement} by removing low-ranked features.
Besides improving the prediction accuracy of the actual models, feature removal has additional benefits for deploying ML models for manufacturing quality prediction.
If the removed features relate to sensor data, their removal also removes the necessity of frequently reading and preprocessing that sensor, which reduces the computational cost of making predictions.
When doing real-time quality prediction to detect potential faults or deviations from the process plan during production, minimizing the time needed to make a prediction and increasing the frequency in which predictions can be made is crucial.

In cases where the model development for quality prediction is considered during the prototyping phase and construction of the manufacturing machine, the relevance of features can inform the selection of physical sensors to be deployed on the machine.
The prototype machine is equipped with a larger set of sensors, and only after evaluating the prediction models is the final set of relevant sensors determined.

Finally, besides the direct relation to the manufacturing case study, we see the benefit of deploying smaller, less complex, and ideally interpretable models \cite{Breiman2001a,Rudin2022}. At the same time, there is a trade-off between simplicity and accuracy, referred to as the Occam dilemma \cite{Breiman2001a}. The simpler a model is made, the less accurate it gets. We see this in the case study from the error difference between the simpler decision tree vs. the more complex gradient-boosting trees or random forest.
By applying explainability methods to reduce the feature space, we again reduce the model complexity, making the final model more interpretable.

\section{\uppercase{Conclusion}}
\label{sec:conclu}
This study showcases the potential of combining ML and explainability techniques to enhance the performance of predictive models of surface quality in the manufacturing sector, specifically in the context of the milling process.
Despite the limitations imposed by data availability, our approach successfully leverages less data-rich ML models, enhancing their efficacy through feature selection based on explainability methods.

For future work, we are interested in extending the application of explainability methods in ML models to other manufacturing processes of our partners beyond milling to create a more comprehensive predictive system.
Additionally, utilizing these ML models as digital twins for the corresponding physical machinery opens new avenues for employing parameter optimization methods.
This integration not only enhances the accuracy of the models but also provides an opportunity for real-time fine-tuning of machine operations, thereby potentially improving efficiency and reducing costs.

\section*{Acknowledgments}
This work is funded by the European Union under grant agreement number 101091783 (MARS Project) and as part of the Horizon Europe HORIZON-CL4-2022-TWIN-TRANSITION-01-03.
\bibliographystyle{apalike}
{\small
\bibliography{example}}

\begin{thebibliography}{}

\bibitem[Bento et~al., 2021]{bento2021improving}
Bento, V., Kohler, M., Diaz, P., Mendoza, L., and Pacheco, M.~A. (2021).
\newblock Improving deep learning performance by using explainable artificial intelligence (xai) approaches.
\newblock {\em Discover Artificial Intelligence}, 1:1--11.

\bibitem[Bins and Draper, 2001]{bins2001feature}
Bins, J. and Draper, B.~A. (2001).
\newblock Feature selection from huge feature sets.
\newblock In {\em Proceedings Eighth IEEE International Conference on Computer Vision. ICCV 2001}, volume~2, pages 159--165. IEEE.

\bibitem[Breiman, 2001]{Breiman2001a}
Breiman, L. (2001).
\newblock Statistical {{Modeling}}: {{The Two Cultures}} (with comments and a rejoinder by the author).
\newblock {\em Statistical Science}, 16(3):199--231.

\bibitem[Brusa et~al., 2023]{Brusa23}
Brusa, E., Cibrario, L., Delprete, C., and Di~Maggio, L. (2023).
\newblock Explainable ai for machine fault diagnosis: Understanding features’ contribution in machine learning models for industrial condition monitoring.
\newblock {\em Applied Sciences}, 13(4):20--38.

\bibitem[Chouhad et~al., 2021]{Chouhad21}
Chouhad, H., Mansori, M.~E., Knoblauch, R., and Corleto, C. (2021).
\newblock Smart data driven defect detection method for surface quality control in manufacturing.
\newblock {\em Meas. Sci. Technol.}, 32(105403):16pp.

\bibitem[Fertig et~al., 2022]{fertig2022machine}
Fertig, A., Weigold, M., and Chen, Y. (2022).
\newblock Machine learning based quality prediction for milling processes using internal machine tool data.
\newblock {\em Advances in Industrial and Manufacturing Engineering}, 4:100074.

\bibitem[Huang et~al., 2016]{huang2016permutation}
Huang, N., Lu, G., and Xu, D. (2016).
\newblock A permutation importance-based feature selection method for short-term electricity load forecasting using random forest.
\newblock {\em Energies}, 9(10):767.

\bibitem[Kummar, 2017]{Kummar17}
Kummar, S.~L. (2017).
\newblock State of the art-intense review on artificial intelligence systems application in process planning and manufacturing.
\newblock {\em Engineering Applications of Artificial Intelligence}, 65:294--329.

\bibitem[Kwon et~al., 2023]{DBLP:journals/cogsr/KwonKC23}
Kwon, H.~J., Koo, H.~I., and Cho, N.~I. (2023).
\newblock Understanding and explaining convolutional neural networks based on inverse approach.
\newblock {\em Cogn. Syst. Res.}, 77:142--152.

\bibitem[Lampathaki et~al., 2021]{Lampathaki21}
Lampathaki, F., Agostinho, C., Glikman, Y., and Sesana, M. (2021).
\newblock Moving from ‘black box’ to ‘glass box’ artificial intelligence in manufacturing with xmanai.
\newblock In {\em 2021 IEEE International Conference on Engineering, Technology and Innovation (ICE/ITMC). Cardiff, UK}.

\bibitem[Lozano-Murcia et~al., 2023]{lozano2023comparison}
Lozano-Murcia, C., Romero, F.~P., Serrano-Guerrero, J., and Olivas, J.~A. (2023).
\newblock A comparison between explainable machine learning methods for classification and regression problems in the actuarial context.
\newblock {\em Mathematics}, 11(14):3088.

\bibitem[Mokhtari~Homami et~al., 2014]{mokhtari2014optimization}
Mokhtari~Homami, R., Fadaei~Tehrani, A., Mirzadeh, H., Movahedi, B., and Azimifar, F. (2014).
\newblock Optimization of turning process using artificial intelligence technology.
\newblock {\em The International Journal of Advanced Manufacturing Technology}, 70:1205--1217.

\bibitem[Mundada and Narala, 2018]{mundada2018optimization}
Mundada, V. and Narala, S. K.~R. (2018).
\newblock Optimization of milling operations using artificial neural networks (ann) and simulated annealing algorithm (saa).
\newblock {\em Materials Today: Proceedings}, 5(2):4971--4985.

\bibitem[Myles et~al., 2004]{myles2004introduction}
Myles, A.~J., Feudale, R.~N., Liu, Y., Woody, N.~A., and Brown, S.~D. (2004).
\newblock An introduction to decision tree modeling.
\newblock {\em Journal of Chemometrics: A Journal of the Chemometrics Society}, 18(6):275--285.

\bibitem[Nguyen and Sakama, 2021]{DBLP:conf/ilp/0002S21}
Nguyen, H.~D. and Sakama, C. (2021).
\newblock Feature learning by least generalization.
\newblock In {\em {ILP}}, volume 13191 of {\em Lecture Notes in Computer Science}, pages 193--202. Springer.

\bibitem[Oreski et~al., 2017]{oreski2017effects}
Oreski, D., Oreski, S., and Klicek, B. (2017).
\newblock Effects of dataset characteristics on the performance of feature selection techniques.
\newblock {\em Applied Soft Computing}, 52:109--119.

\bibitem[Otchere et~al., 2022]{otchere2022application}
Otchere, D.~A., Ganat, T. O.~A., Ojero, J.~O., Tackie-Otoo, B.~N., and Taki, M.~Y. (2022).
\newblock Application of gradient boosting regression model for the evaluation of feature selection techniques in improving reservoir characterisation predictions.
\newblock {\em Journal of Petroleum Science and Engineering}, 208:109244.

\bibitem[Pawar et~al., 2021]{pawar2021modelling}
Pawar, S., Bera, T., and Sangwan, K. (2021).
\newblock Modelling of energy consumption for milling of circular geometry.
\newblock {\em Procedia CIRP}, 98:470--475.

\bibitem[Pedregosa et~al., 2011]{pedregosa2011scikit}
Pedregosa, F., Varoquaux, G., Gramfort, A., Michel, V., Thirion, B., Grisel, O., Blondel, M., Prettenhofer, P., Weiss, R., Dubourg, V., et~al. (2011).
\newblock Scikit-learn: Machine learning in python.
\newblock {\em the Journal of machine Learning research}, 12:2825--2830.

\bibitem[Postel et~al., 2020]{postel2020ensemble}
Postel, M., Bugdayci, B., and Wegener, K. (2020).
\newblock Ensemble transfer learning for refining stability predictions in milling using experimental stability states.
\newblock {\em The International Journal of Advanced Manufacturing Technology}, 107:4123--4139.

\bibitem[Prasad et~al., 2006]{prasad2006newer}
Prasad, A.~M., Iverson, L.~R., and Liaw, A. (2006).
\newblock Newer classification and regression tree techniques: bagging and random forests for ecological prediction.
\newblock {\em Ecosystems}, 9:181--199.

\bibitem[Rudin et~al., 2022]{Rudin2022}
Rudin, C., Chen, C., Chen, Z., Huang, H., Semenova, L., and Zhong, C. (2022).
\newblock Interpretable machine learning: {{Fundamental}} principles and 10 grand challenges.
\newblock {\em Statistics Surveys}, 16:1--85.

\bibitem[Schorr et~al., 2020]{schorr20}
Schorr, S., Moller, M., Heib, J., Fang, S., and Bahre, D. (2020).
\newblock Quality prediction of reamed bores based on process data and machine learning algorithm: A contribution to a more sustainable manufacturing.
\newblock {\em Procedia Manufacturing}, 43:519--526.

\bibitem[Senoner et~al., 2022]{DBLP:journals/mansci/SenonerNF22}
Senoner, J., Netland, T.~H., and Feuerriegel, S. (2022).
\newblock Using explainable artificial intelligence to improve process quality: Evidence from semiconductor manufacturing.
\newblock {\em Manag. Sci.}, 68(8):5704--5723.

\bibitem[Sofianidis et~al., 2021]{sofianidis2021review}
Sofianidis, G., Rožanec, J.~M., Mladenić, D., and Kyriazis, D. (2021).
\newblock A review of explainable artificial intelligence in manufacturing.

\bibitem[Sun et~al., 2022]{sun2022utilizing}
Sun, H., Servadei, L., Feng, H., Stephan, M., Santra, A., and Wille, R. (2022).
\newblock Utilizing explainable ai for improving the performance of neural networks.
\newblock In {\em 2022 21st IEEE International Conference on Machine Learning and Applications (ICMLA)}, pages 1775--1782. IEEE.

\bibitem[Sundararajan and Najmi, 2020]{sundararajan2020many}
Sundararajan, M. and Najmi, A. (2020).
\newblock The many shapley values for model explanation.
\newblock In {\em International conference on machine learning}, pages 9269--9278. PMLR.

\bibitem[Venkatesh and Anuradha, 2019]{venkatesh2019review}
Venkatesh, B. and Anuradha, J. (2019).
\newblock A review of feature selection and its methods.
\newblock {\em Cybernetics and information technologies}, 19(1):3--26.

\bibitem[Yoo and Kang, 2021]{Yoo21}
Yoo, S. and Kang, N. (2021).
\newblock Explainable artificial intelligence for manufacturing cost estimation and machining feature visualization.
\newblock {\em Expert Systems with Applications}, 183.

\end{thebibliography}



\end{document}